\documentclass{article}
\usepackage{spconf,amsmath,graphicx}

\usepackage{algorithm}
\usepackage{algorithmic}

\usepackage{threeparttable}
\usepackage{float}
\usepackage{colortbl}  
\usepackage{xcolor}
\usepackage{array}   

\usepackage{times}
\usepackage{epsfig}
\usepackage{graphicx}
\usepackage{amsmath}
\usepackage{amssymb}
\usepackage{multirow}

\usepackage{newfloat}
\usepackage{listings}

\usepackage{hyperref}
\usepackage{marvosym}

\title{Hyneter: Hybrid Network Transformer for Object Detection}
\name{Dong Chen$^{\href{mailto:1910691@tongji.edu.cn}{\textrm{\Letter}}}$	$ $ \qquad Duoqian Miao$^{\ast}$ \thanks{*Corresponding author} $ $ \qquad Xuerong Zhao$ $}

\address{ Department of Computer Science and Technology, Tongji University \\}

\begin{document}
	
	\maketitle
	
	\begin{abstract}
		In this paper, we point out that the essential differences between CNN-based and Transformer-based detectors, which cause worse performance of small object in Transformer-based methods, are the gap between local information and global dependencies in feature extraction and propagation. To address these differences, we propose a new vision Transformer, called \textbf{Hy}brid \textbf{Net}work Transform\textbf{er} (Hyneter), after pre-experiments that indicate the gap causes CNN-based and Transformer-based methods to increase size-different objects results unevenly. Different from the divide and conquer strategy in previous methods, Hyneters consist of Hybrid Network Backbone (HNB) and Dual Switching module (DS), which integrate local information and global dependencies, and transfer them simultaneously. Based on the balance strategy, HNB extends the range of local information by embedding convolution layers into Transformer blocks, and DS adjusts excessive reliance on global dependencies outside the patch. Ablation studies illustrate that Hyneters achieve the state-of-the-art performance by a large margin of $+2.1 \sim 13.2 AP$ on COCO, and $+3.1 \sim 6.5 mIoU$ on Visdrone with lighter model sizes in object detection. Furthermore,  Hyneters achieve the state-of-the-art results on multiple vision tasks, such as object detection ($60.1 AP$ on COCO, $46.1$ on Visdrone), semantic segmentation ($54.3 AP$ on ADE20K), and instance segmentation ($48.5 AP^{mask}$ on COCO), and surpass previous best methods. 
	\end{abstract}
%
	
	\section{Introduction}
	
	Convolutional neural networks (CNNs) have dominated computer vision modeling for years.  With the help of increasingly large neural networks and progressively complex convolution structures, the performance has seen significant improvement in recent time. However, scholars have focused on greater model size, more diverse convolution kernel, and more sophisticated structures of network, which lead to a less progress of general performance with disproportionate huge model size.
	
	On the other hand, Transformer has made tremendous progress in vision tasks, which originates from natural language processing (NLP). Designed for sequence modeling and transduction tasks, the Transformer is notable for its use of attention to model global dependencies in the feature. The tremendous success in NLP has led researchers to investigate its adaptation to computer vision, where it has recently demonstrated promising results on certain tasks. Compared to CNN-based methods, vision Transformer and its follow-ups (including hybrid methods) expose the difference in size-sensitive performance, for they adopt different strategies for local information and global dependencies \cite{dong2021attention}.
	
	The essential differences between CNN-based and Transformer-based detectors are derived from the gap between local information and global dependencies in feature extraction and propagation. However, we have not found enough  studies on these differences. In this paper, we devote to find the answer and propose a new vision Transformer.
	
	\begin{figure}[h]
		\centering
		\includegraphics[scale= 0.18]{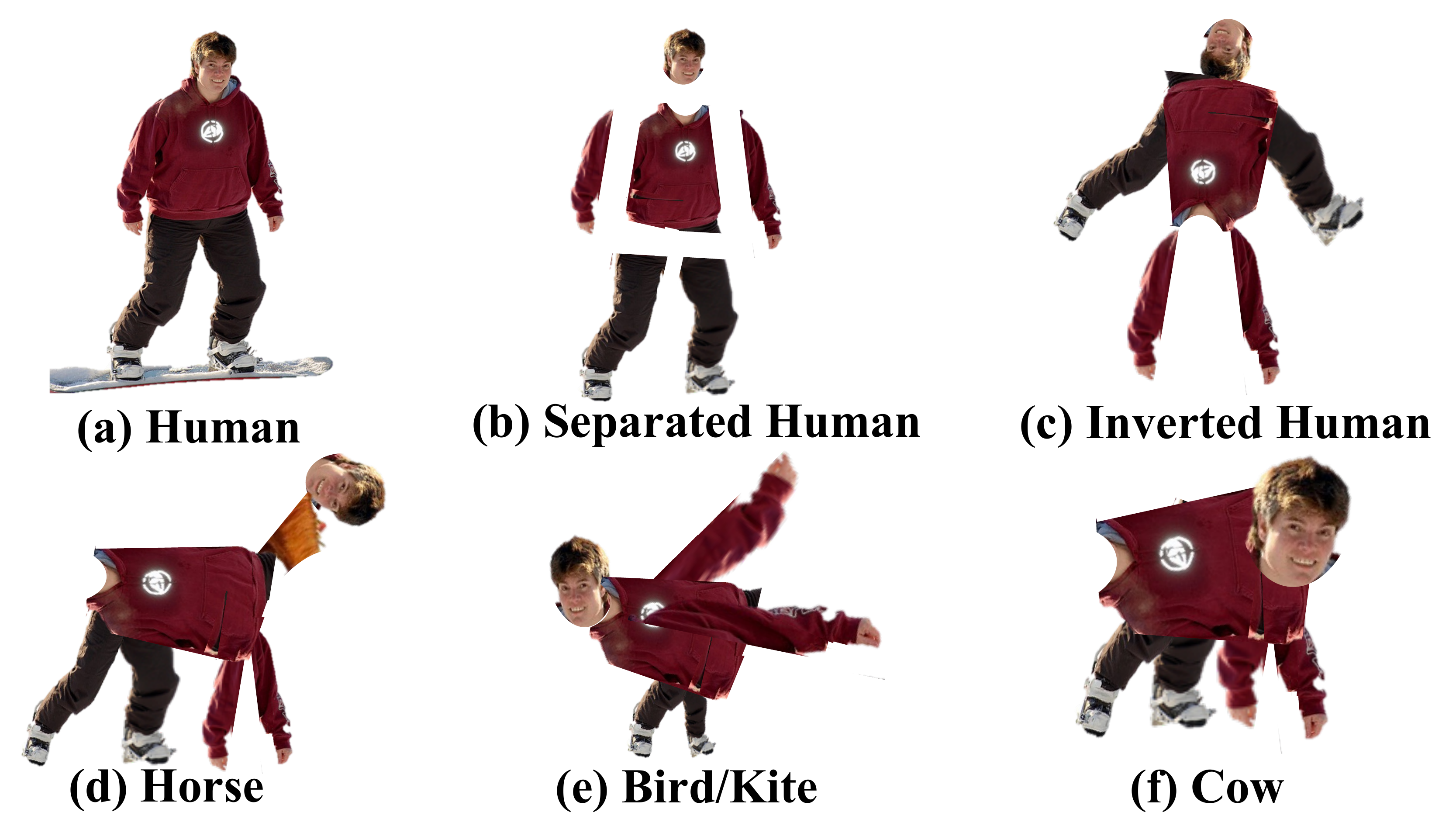} 
		\caption{An illustration of restructured objects. We restructure thousands of objects in multiple-class images of COCO and Visdrone. For example, there are 3 objects ($(d) \sim (f)$) supposed to detected as \textit{unrecognized labels}, but as  \textit{Pseudo labels} (horse, bird/kite, and cow) by Transformer-based detectors. Transformer-based detectors should detect $(b) and (c)$ as \textit{unrecognized labels}, but \textit{True label} (human).} 
		\label{fig:reorg} 
	\end{figure}
	
	
	The exploration begins with an unexpected experiment shown in Figure \ref{fig:reorg}. We restructure thousands of objects in multiple-class images with diverse backgrounds such as ocean, grassland, sky, indoor environment, snow, playground, desert, forest, etc. A human, for example, is restructured as horse, bird/kite, cow, etc. In Figure \ref{fig:reorg}, $(d) \sim (f)$ are supposed to detected as \textit{unrecognized labels}, but as  \textit{Pseudo labels} (horse, bird/kite, and cow) by Transformer-based detectors. However, CNN-based detectors show much better performance. This rate of being detected as \textit{pseudo labels} (\textit{Pseudo Rate}) demonstrates that Transformer-based methods are reliant on global dependencies and obtain inadequate local information of feature in details \cite{dong2021attention}. However, the CNN-based methods are just the opposite.
	
	\begin{figure}[h]
		\centering
		\includegraphics[scale= 0.20]{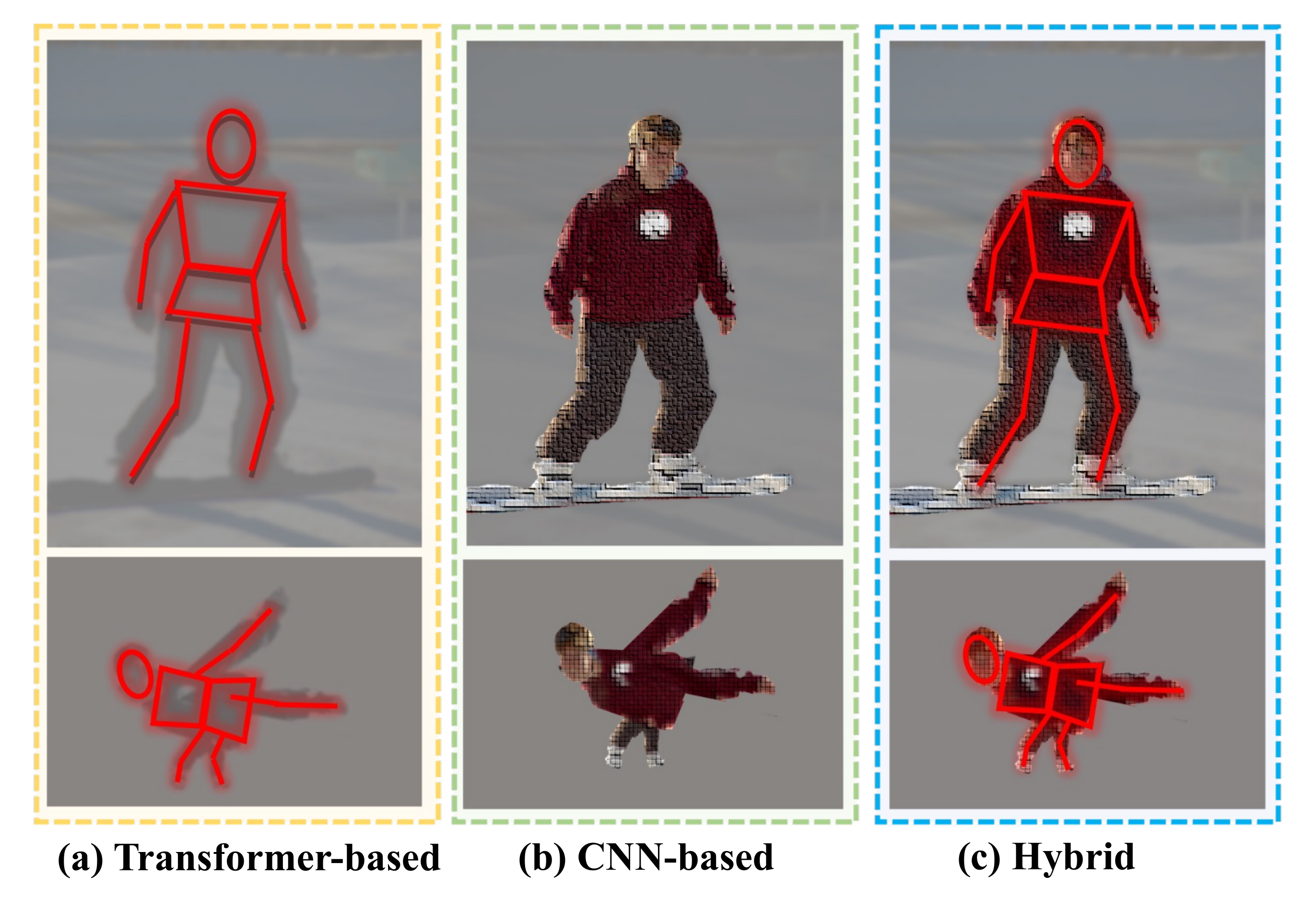} 
		\caption{An illustration of feature maps on Transformer-based, CNN-based and Hybrid methods. Hybrid feature map (c) integrates the characteristics of global dependencies (a) and local information (b), which is beneficial to objects of all sizes.} 
		\label{fig:views} 
	\end{figure}
	
	
	The CNN-based methods extract feature with rich local information by convolution layers \cite{chen2021you,  raghu2021vision}. While Transformer-based methods extract feature by providing the capability to decode and encode global dependencies in Transformer blocks \cite{li2022exploring, dai2021dynamic} (see Figure \ref{fig:views}). Compared to CNN-based methods, Transformer-based methods have worse  performance in small objects (see Table \ref{tab:obco2} and \ref{tab:obco3}).
	
	In this paper, we demonstrate that the essential difference between CNN-based and Transformer-based detectors is the gap between local information and global dependencies in feature extraction and propagation. Firstly, We screen 4 influence factors: the number of CNN layer (CL), the number of Transformer block (TB), the number of token (NT), and attention score scaler ($\delta$). Pre-experiments are conducted on COCO object detection about the influence of 4 factors on evaluation criterion($AP$, $AP_{S}$, \textit{Pseudo Rate}). Then, the pre-experiments indicate local information tends to help improve $AP$ by increasing $AP_{S}$, and global dependencies tend to achieve the same effect by increasing $AP_{M}$ and $AP_{L}$,  which cause the essential difference between CNN-based and Transformer-based detectors. Meanwhile, both of them will interfere with each other (see Tables \ref{tab:cnnlayer} $\sim$ \ref{tab:pearson}).
	
	Given the above conclusions, we propose a new vision Transformer, called \textbf{Hy}brid \textbf{Net}work Transform\textbf{er} (Hyneter), which consists of Hybrid Network Backbone (HNB) and Dual Switching module (DS). Hybrid network backbone is presented with equivalent position of intertwined distribution of convolution and self-attention. Our backbone extends the range of local information by embedding convolution layers into Transformer blocks in stages, so that local information and global dependencies will be passed to \textit{Neck} or \textit{Head} simultaneously. The Dual Switching module establishes cross-window connections in order to maintain local information inside the patch, while weakening excessive reliance on global dependencies outside the patch. Based on the balance strategy, Hyneters integrate and transfer local information and global dependencies simultaneously, so they are able to significantly improve performance.  
	
	Ablation studies illustrate that Hyneters with HNB and DS achieve the state-of-the-art performance by a large margin of $+2.1 \sim 13.2 AP$ on COCO, and $+3.1 \sim 6.5  mIoU$ on Visdrone in object detection. Furthermore,  Hyneters achieve the state-of-the-art performance on multiple tasks significantly, such as object detection($60.1 AP$ on COCO, $46.1$ on Visdrone),  semantic segmentation ($54.3 AP$ on ADE20K), and instance segmentation ($48.5 AP^{mask}$ on COCO), and surpass previous best methods (see Table \ref{tab:abla} $\sim$ \ref{tab:ssADK}).
	
	\section{Related Work}
	
	\textbf{CNN based vision backbones.} The backbone networks of deep learning are evolving. LeNet (1998)\cite{1998Gradient},
	AlexNet (2012)\cite{krizhevsky2017imagenet}, VGGNet (2014)\cite{simonyan2014very},
	GoogLeNet (2014)\cite{2014Going}, ResNet (2015)\cite{2016Deep}, and MobileNet (2017)\cite{howard2017mobilenets}
	are preserved in development of deep learning .
	EfficientNet (2019)\cite{tan2019efficientnet} proposes a more generalized idea on the optimization of current
	classification networks, arguing that the three common ways of enhancing network metrics,
	namely widening the network, deepening the network and increasing the resolution,
	should not be independent of each other.
	
	Along with the backbone evolving, the convolutional kernels \cite{fu2017dssd,noh2015learning,radford2015unsupervised} are also changing.
	Deformable conv\cite{dai2017deformable,2019Deformable} adds an offset variable to the position of each sampled point in the convolution kernel,
	enabling random sampling around the current position without being restricted to the previous regular grid points.
	Dilated conv\cite{chen2017rethinking,2016Multi} can effectively focus on the semantic information of the local pixel blocks,
	instead of letting each pixel rub together with the surrounding blocks, which affects the detail of segmentation.
	
	\textbf{Transformer based vision backbones.} The pioneering work of ViT \cite{dosovitskiy2020image} directly applies a Transformer architecture on non-overlapping image patches for image classification. ViT and its follow-ups \cite{touvron2021training, yuan2021tokens, han2021Transformer, wang2021pyramid} achieve an impressive speed-accuracy trade-off on image classification compared to convolutional networks. The results of ViT on image classification are encouraging, but its architecture is unsuitable for use as a general-purpose backbone network on dense vision tasks or when the input image resolution is high, due to its low-resolution feature maps and the quadratic increase in complexity with image size. 
	
	DETR \cite{carion2020end} and Swin Transformer \cite{liu2021swin}, following ViT and variants, are representative methods in computer vision. DETR and its follow-ups (UP-DETR\cite{dai2021up}, Conditional DETR\cite{meng2021conditional}, OW-DETR \cite{gupta2022ow}, Deformable DETR \cite{zhu2020deformable}) demonstrate excellent plasticity and flexibility in computer vision tasks. Meanwhile, Swin Transformer is both efficient and effective, achieving state-of-the-art accuracy on both object detection and semantic segmentation. 
	
	\textbf{Hybrid network vision backbones.} Many hybrid backbones \cite{srinivas2021bottleneck} are presented in previous works, which put convolution and self-attention in the non equivalent position. Previous methods employ self-attention within the CNN backbone architecture or use them outside the CNN backbone architecture. Furthermore, representative hybrid methods completely cleavage the relation of local information and global dependencies by separated distribution of convolution and self-attention. 
	
	Different from pure attention models (such as SASA \cite{ramachandran2019stand}, LRNet \cite{hu2019local}, SANet \cite{zhao2020exploring}, Axial-SASA \cite{wang2020axial, ho2019axial} and ViT), VideoBERT \cite{sun2019videobert}, VILBERT \cite{lu2019vilbert}, CCNet \cite{huang2019ccnet} employ self-attention on the top of backbone architecture. AA-ResNet \cite{bello2019attention} also attempted to replace a fraction of spatial convolution channels with self-attention. But hybrid network methods proved to be imbalanced in size-sensitive performance, as they utilize local information and global dependencies unequally.
	
	
	\section{Analysis of Influence Factors}
	
	In this section, pre-experiments are conducted to analyze the qualitative and quantitative influence of 4 factors to local information and global dependencies in Transformer-based detectors. The comparisons will provide foundations to create a new Transformer, Hyneter.
	
	Without loss of generality, the representative DETR is adopted to dissect the influence of CNN layers (CL), Transformer blocks (TB), the number of token (NT) and and attention score scaler ($\delta$)  \cite{vaswani2017attention}. The $\delta$ is a parameter that controls the calculation of attention score in Transformer:
	
	\begin{equation}
		\text { attention score }=\left\{\begin{array}{l}
			q_{i} \cdot k_{l}, i=l \\
			\delta q_{i} \cdot k_{l}, i \neq l
		\end{array}\right.
	\end{equation}
	
	\begin{figure}[h]
		\centering
		\includegraphics[scale= 0.33]{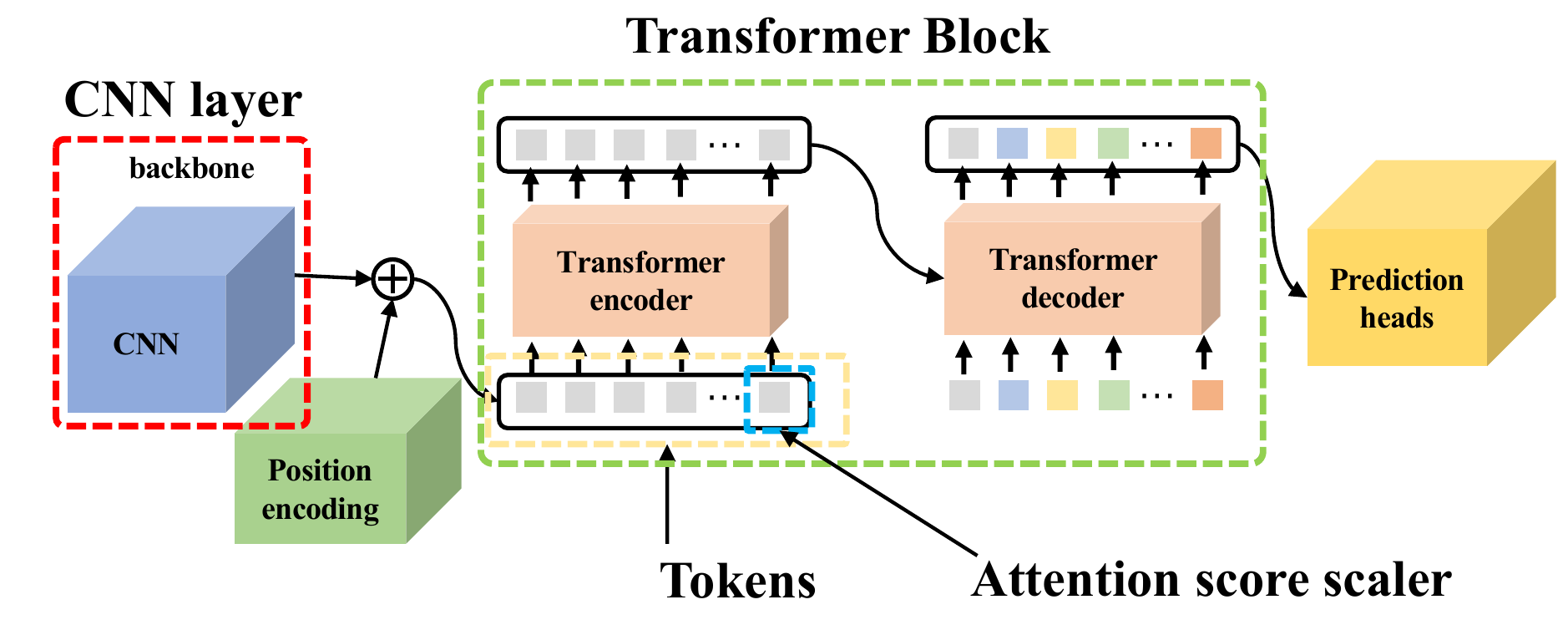} 
		\caption{An illustration of DETR stressing on 4 factors: CNN layer, Transformer block, tokens, and attention score scaler.} 
		\label{fig:detr} 
	\end{figure}
	
	\vspace{-0.2cm}
	
	\begin{table}[h]\small

		\setlength{\tabcolsep}{0.8mm}{
			\begin{tabular}{|l|c|c|c|c|c|c|c|c|}
				\hline
				Backbone   & \#param. & $AP$   & $AP_{S}$  & $AP$/$AP_{S}$ & Pseudo & True & Unre \\ \hline
				R-34  & 27M      & 38.6 & 18.6 & 2.08   & 67.0   & 30.0 & 3.0  \\ \hline
				R-50  & 41M      & 42.0 & 20.5 & 2.05   & 60.4   & 33.9 & 5.7  \\ \hline
				R50-DC5    & 41M      & 43.3 & 22.5 & 1.92   & 55.7   & 36.1 & 8.2  \\ \hline
				R-101 & 60M      & 43.5 & 21.9 & 1.99   & 51.2   & 38.4 & 10.4 \\ \hline
				R101-DC5   & 60M      & 44.9 & 23.7 & 1.89   & 42.6   & 45.2 & 12.2 \\ \hline
				R-152 & 92M      & 45.4 & 24.3 & 1.86   & 40.0   & 47.7 & 12.3 \\ \hline
		\end{tabular}}
		\caption{Comparison(\%) on DETR with ResNet-X backbones on COCO val set. We train DETR with setting as Technical details in \cite{carion2020end}. R50-DC5 means ResNet-50 with dilated C5 stage. \textit{Pseudo} means \textit{Pseudo Rate}; \textit{True} means true label rate; \textit{Unre} means unrecognized label rate.}
		\label{tab:cnnlayer}
	\end{table}
	
	\begin{table}[h]\small

		\setlength{\tabcolsep}{1.1mm}{
			\begin{tabular}{|c|c|c|c|c|c|c|c|c|}
				\hline
				Blocks   & \#params & $AP$   & $AP_{S}$  & $AP$/$AP_{S}$ & Pseudo & True & Unre \\ \hline
				$\times1.0$              & 41M      & 42.0 & 20.5 & 2.05   & 60.4   & 33.9 & 5.7  \\ \hline
				$\times2.0$             & 51M      & 44.1 & 21.0 & 2.10   & 66.5   & 28.7 & 4.8  \\ \hline
				$\times3.0$              & 61M      & 45.4 & 20.4 & 2.23   & 68.7   & 25.4 & 5.9  \\ \hline
				$\times4.0$              & 70M      & 46.1 & 20.3 & 2.27   & 72.9   & 23.1 & 4.0  \\ \hline
		\end{tabular}}
		\caption{Comparison(\%) on DETR with variant Transformer blocks and ResNet-50 backbone on COCO val set. $\times3.0$ means that DETR with 3 Transformer blocks and ResNet-50.}
		\label{tab:transbloack}
	\end{table}
	
	\begin{table}[h]\small

		\setlength{\tabcolsep}{1.1mm}{
			\begin{tabular}{|c|c|c|c|c|c|c|c|c|}
				\hline
				Tokens   & \#params & $AP$   & $AP_{S}$  & $AP$/$AP_{S}$ & Pseudo & True & Unre \\ \hline
				$\times1.0$   & 41M      & 42.0 & 20.5 & 2.05   & 60.4   & 33.9 & 5.7  \\ \hline
				$\times1.5$   & 62M      & 43.5 & 20.8 & 2.10   & 58.5   & 35.7 & 5.8  \\ \hline
				$\times2.0$   & 83M      & 44.1 & 21.2 & 2.09   & 57.0   & 36.4 & 6.6  \\ \hline
				$\times2.5$   & 103M     & 45.0 & 21.6 & 2.08   & 55.4   & 37.1 & 7.5  \\ \hline
		\end{tabular}}
		\caption{Comparison(\%) on DETR with variant number of tokens and ResNet-50 backbone on COCO val set. $\times2.0$ means that DETR with $2.0\times HW$ tokens and ResNet-50.}
		\label{tab:token}
	\end{table}
	
	\begin{table}[h]\small

		\setlength{\tabcolsep}{1.1mm}{
			\begin{tabular}{|c|c|c|c|c|c|c|c|c|}
				\hline
				Scalers   & \#params & $AP$   & $AP_{S}$  & $AP$/$AP_{S}$ & Pseudo & True & Unre \\ \hline
				$\times1.0$   & 41M      & 42.0 & 20.5 & 2.05   & 60.4   & 33.9 & 5.7  \\ \hline
				$\times1.5$   & 41M      & 42.9 & 20.6 & 2.06   & 65.0   & 32.5 & 2.5  \\ \hline
				$\times2.0$   & 41M      & 43.5 & 21.0 & 2.07   & 67.0   & 30.7 & 2.3  \\ \hline
				$\times2.5$   & 41M      & 44.7 & 21.5 & 2.08   & 71.1   & 27.0 & 1.9  \\ \hline		
		\end{tabular}}
		\caption{Comparison(\%) on DETR with variant attention score scaler and ResNet-50 backbone on COCO val set. $\times2.0$ means that DETR with $2.0\times  attention$ $score $ to other $score = q_{i} \cdot k_{l \neq i} $ and ResNet-50.}
		\label{tab:scaler}
	\end{table}
	
	\begin{table}[h]\small

		\setlength{\tabcolsep}{3.4mm}{
			\begin{tabular}{|c|c|c|c|c|}
				\hline
				$\rho$ & \multicolumn{1}{c|}{$AP$} & \multicolumn{1}{c|}{$AP_{s}$} & \multicolumn{1}{c|}{$AP/AP_{s}$} & \multicolumn{1}{c|}{Pseudo} \\ \hline
				CNN layers             &   0.92 &0.98& \cellcolor{gray}-0.98 &\cellcolor{lightgray}-0.92       \\ \hline
				Trans blocks           &    0.99 &\cellcolor{lightgray}-0.50& 0.98& 0.98      \\ \hline
				Tokens                 &   0.98& 1.00 &0.48 &\cellcolor{gray}-1.00       \\ \hline
				Scaler $\delta$            &    0.99 &0.75& 1.00 &0.99       \\ \hline
		\end{tabular}}
		\caption{Pearon correlation coefficinet ($\rho$) comparison(\%) on factors and evaluating indicators ($AP$, $AP_{S}$, $AP/AP_{S}$, and Pseudo). Gray indicates negative correlation and white indicates positive correlation.}
		\label{tab:pearson}
	\end{table}
	
	As shown in Table \ref{tab:cnnlayer}, \ref{tab:transbloack}, \ref{tab:token}, and \ref{tab:scaler}, comparisons on DETRs with different factors demonstrate the qualitative and quantitative relationship among local information, global dependencies, 4 factors and detector performance. With the help of Pearon correlation coefficinet ($\rho$) in Table \ref{tab:pearson}, the rules are summarized as follows:
	
	\begin{itemize}
		\item With the increase of CNN layers, the detectors will pay more attention to local information, reducing the reliance on global dependencies, and will gradually improve $AP$ and $AP_{S}$.
		\item The increase of Transformer blocks will promote detectors to rely more on global dependencies, thereby improving performance, but hurt $AP_{S}$.
		\item The increase of token will weaken the methods' reliance on global dependencies, and simultaneously improve the methods $AP$ and $AP_{S}$, but increase model size.
		\item The increase of the attention score scaler simultaneously improves $AP$, increasing the reliance on global dependencies, nor does it increase model size.
	\end{itemize}
	
	\textbf{Conclusion.} Local information tends to increase $AP_{S}$ to improve $AP$, and global dependencies tends to increase $AP_{M}$ and $AP_{L}$ in order to improve $AP$. Meanwhile, both of them will interfere with each other. The gap between local information and global dependencies in feature extraction and propagation causes worse performance of small object in Transformer-based methods. Computing self-attention impedes extraction of local information feature, while convolution layers stop extracting feature of global dependencies (see (a) and (b) in Figure \ref{fig:views}).
	
	
	\section{Hybrid Network Transformer}
	
	In view of the above conclusions, we propose a new vision Transformer, called \textit{Hybrid Network Transformer}, that capably serves as a backbone for multiple computer vision tasks, which consists of Hybrid Network Backbone and Dual Switching module.
	
	\begin{figure*}[h]
		\centering
		\includegraphics[scale= 0.55]{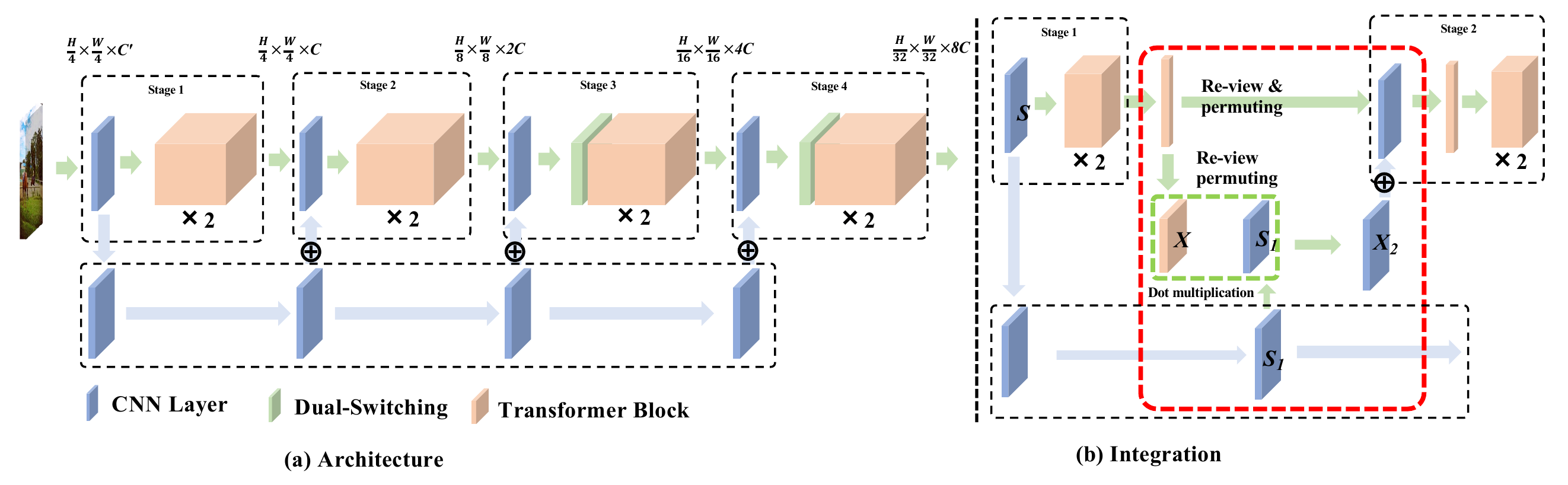} 
		\caption{(a) The architecture of Hyneter 1.0.  There are 2 Transformer blocks in one stage of Transformer blocks (top) and  2-layer multi-granularity convolution layers in one stage of CNN layers part (bottom). Positional encoding, patch partition, and self-attention in the first Transformer block, but patch partition and self-attention in others. (b) An illustration of unidirectional feature integration between Transformer block (top) and CNN layer (bottom), as illustrated in Figure \ref{fig:views} (c).} 
		\label{fig:hnd} 
	\end{figure*}
	
	An overview of the \textbf{Hy}brid \textbf{Net}work Transform\textbf{er} architecture (Hyneter) is presented in Figure \ref{fig:hnd} (a), which illustrates the basic version. Data is preprocessed as Method in \cite{liu2021swin}.
	
	\subsection{Hybrid Network Backbone}
	
	Many hybrid backbones \cite{srinivas2021bottleneck} are presented in previous works, which put convolution and self-attention in the non equivalent position. Previous methods employ self-attention within the CNN backbone architecture or use them outside. Furthermore, representative hybrid methods (such as DETR, see Figure \ref{fig:detr}) completely cleavage the relation of local information and global dependencies by separated distribution of convolution and self-attention. Hybrid network backbone is presented with equivalent position of intertwined distribution of convolution and self-attention. Our backbone extends the range of local information, so that local information and global dependencies will be passed to \textit{Neck} or \textit{Head} simultaneously.
	
	There are 4 stages in our backbone, starting with a convolution layer of 3 multi-granularity kernels. The number of tokens is reduced by this multi-granularity convolution layer, and dimension is multiplied.  The data feature $S$ ($C^{\prime} \times \frac{H}{4} \times \frac{W}{4}$) will be sent into convolution layers and Transformer blocks. 
	
	As shown in Figure \ref{fig:hnd} (b), the Transformer blocks extract feature maps of global dependencies and CNN layers extract feature maps of local information in the Stage 1 and 2. The output ($C \times \frac{H \times W}{4 \times 4}$) of the final Transformer block in Stage 1 will be re-viewed and  permuted as $X$($C \times \frac{H}{4} \times \frac{W}{4}$). After the convolution layers, the $S$ turns into $S_{1}$ with the same size ($C \times \frac{H}{4} \times \frac{W}{4}$). The dot product between $S_{1}$ and $X$ is the key operation of combination for global dependencies and local information. The $X_{1}$ ($X_{1} = S_{1} \cdot X$) after dot product operation, will go to activation function $X_{2} = tanh(X_{1})$. The addition of $X_{2}$ and $X$ copy will be the output of Stage 1. After being re-viewed and  permuted twice, the addition turns to the input ($X^{\prime}$) of Stage 2.
	
	With hybrid network approach, consecutive self-attention Transformer blocks are computed as
	
	\begin{equation}
		\begin{aligned}
			&X=\operatorname{Re-view}(\text{ GMSA }(S)) \\
			&S_{1}=\operatorname{Conv}_{1}(S) \oplus \operatorname{Conv}_{2}(S) \oplus \operatorname{Conv}_{3}(S) \\
			&X_{2}=\tanh \left(X \cdot S_{1}\right) \\
			&X^{\prime}=\operatorname{Re-view}(X \oplus X_{2}
			)
		\end{aligned}
	\end{equation}
	
	The Transformer blocks in Stage 1 and 2 are pure self-attention with maintaining the number of tokens, and together with interfaces for convolution layer output. The blocks in Stage 3 and 4 will be implemented with Dual Switching. GMSA means global multi-head self-attention.
	
	\subsection{Dual Switching}
	
	The Dual Switching module will be implemented in Stage 3 and 4, in order to maintain local information while weakening excessive reliance on global dependencies.
	
	Global dependencies from global self-attention are conducted in Transformer blocks, where the dependencies among tokens are computed. With resepect to NT, the computation results in quadratic complexity, which is inadequate for many vision tasks with huge NT. For efficiency, the global multi-head self-attention (GMSA) will be implemented within local windows in a non-overlapping manner.
	
\begin{figure}[h]
		\centering
		\includegraphics[scale= 0.38]{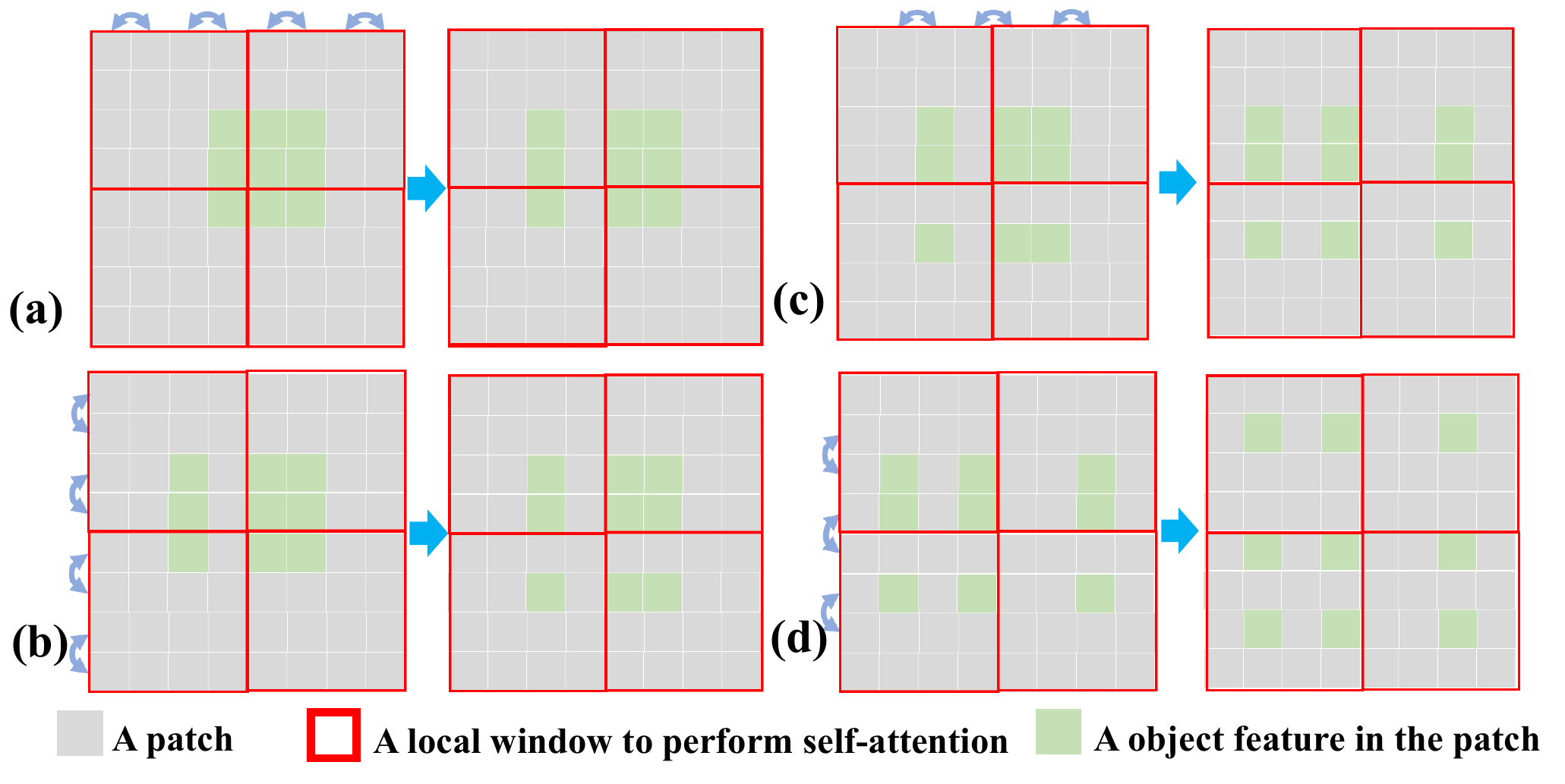} 
		\caption{An illustration of Dual Switching. The process is implementing as (a)$\rightarrow$(b)$\rightarrow$(c)$\rightarrow$(d).} 
		\label{fig:dsm} 
	\end{figure}

	As illustrated in Figure \ref{fig:dsm}, the ouput of Transformer block will be re-viewed and permuted as $X$($C \times \frac{H}{4} \times \frac{W}{4}$). Then, adjacent columns in the feature map will switch with each other. After the column switching, adjacent rows in the feature map will switch with each other, too. The solo-switching is finished. Finally, the interlaced columns/rows in solo-switched feature map will switch with each other, again. 
	
	The Dual Switching module establishes cross-window connections while maintaining local information in the patch, which is followed by layerNorms (LN), Transformer blocks, and multi-layer perceptions (MLP) with residual connection modules.  
	
	After Stage 1 and 2 in our backbone, the feature in a patch with abundant local information has established  considerable global dependencies with surrounding patches. Dual Switching suspends the procedure of establishing excessive global dependencies, meanwhile, retaining local information for small object performance ($AP_{S}$). With Dual Switching module, the process is computed as
	
	\begin{equation}
		\begin{aligned}
			&X_{l}=\text { Dual-Switch }\left(X_{l}\right) \\
			&X_{l+1}=\text{ GMSA }\left(\operatorname{LN}\left(X_{l}\right)\right)+X_{l} \\
			&X_{l+1}^{\prime}=\operatorname{MLP}\left(\operatorname{LN}\left(X_{l+1}\right)\right)+X_{l+1}
		\end{aligned}
	\end{equation}
	
	where $X_{l}$ and $X_{l+1}^{\prime}$ denote the the feature in Stage $l$ and the input of Stage $l+1$.
	
	\vspace{-0.2cm} 
	
	\subsection{Architecture Variants}
	
	We establish basic model, called Hyneter 1.0, to have of size and computation complexity  similar to DETR-DC5-R101. This paper also presents Hyneter Plus and Hyneter Max, which are 2 versions of around $2.0 \times$ and $4.0 \times$ the model size and computation complexity, respectively. The architecture hyper-parameters of these model variants are:
	
	\begin{itemize}
		\item Hyneter 1.0 : $d =96$, CNN layer = \{2, 2, 2, 2\}, Transformer block = \{2, 2, 2, 2\}
		\item Hyneter Plus : $d =96$, CNN layer = \{2, 2, 3, 2\}, Transformer block = \{2, 2, 6, 2\}
		\item Hyneter Max : $d =128$, CNN layer = \{2, 2, 6, 2\}, Transformer block = \{2, 2, 18, 2\}
	\end{itemize}
	
	where $d$ is the channel number of the Transformer block in the first stage.
	
	\section{Experiments}
	
	In this section, we conduct experiments on multiple datasets in several vision tasks. In the following, we first ablate the important design elements of Hyneter. Then, we compare the proposed Hyneter architecture with the previous state-of-the-arts on the three tasks.
	
	\subsection{Ablation studies}
	
	\textbf{Settings.} The following experiments were conducted on MS COCO 2017 dataset using two GeForce RTX 3090 GPUs and 2 Tesla V100 PCIe 32GB GPUs. All models under pytorch framework are standard models without using any tricks. For the ablation study and comparisons, we consider four typical object detection frameworks: Swin Transformer (V1, V2)\cite{liu2021swin, liu2022swin}, and DETRs (DETR\cite{carion2020end}, UP-DETR\cite{dai2021up}, Conditional DETR\cite{meng2021conditional}).
	
	\textbf{Dataset.} We perform experiments on COCO 2017 detection datasets, containing 118k training images, 5k validation images and 20K test-dev images. The ablation study is performed using the validation set, and a system-level comparison is reported on test-dev. Each image is annotated with bounding boxes and panoptic segmentation. There are 7 instances per image on average, up to 63 instances in a single image in training set, ranging from small to large on the same images.
	
	\textbf{Training.}  Hyneter is trained with Adamw\cite{loshchilov2017decoupled} and SGD optimizers, changing Adamw to SGD until very final stage. We adopt Hyneter models with the learning rate ($2^{-5}$) for backbone. The backbone is the ImageNet-pretrained model with batchnorm layers fixed, and the Transformer parameters are initialized using the Xavier initialization scheme. The weight decay is set to be $10^{-4}$.
	
	We conduct ablation studies on COCO 2017 object detection. Table \ref{tab:abla} list the results of Hyneter variants with Mask R-CNN. Our architecture with Hybrid network backbone (HNB) or Dual Switching (DS) brings consistent $+3.2 \sim 4.8$ $AP$ and $+4.1 \sim 6.8$ $AP_{S}$ gains over pure Transformer detectors. Furthermore, HNB brings $+ 1.6\sim 2.7 $ $AP$ and $+ 1.7 \sim 3.8 $ $AP_{S}$ gains over original detectors, just with slightly larger model size. Meanwhile, DS gets $+ 1.6\sim 2.1 $ $AP$ and $+ 1.2 \sim 3.0 $ $AP_{S}$ gains over original detectors, with the same model size.
	
	\begin{table}[h]\small

		\setlength{\tabcolsep}{1.0mm}{
			\begin{tabular}{|l|c|c|c|c|c|c|c|}
				\hline
				Method                        & Originals & HNB & DS & $AP$   & $AP_{s}$  & $AP/AP_{s}$ & \#param. \\ \hline
				\multicolumn{8}{|c|}{Hyneter 1.0}                                                                                                         \\ \hline
				\multirow{3}{*}{baseline}  & \checkmark          &     &    & 52.3 & 21.5 & 2.43   &    85M      \\ 
				& \checkmark          & \checkmark    &    & 55.0 & 25.3 & 2.17   &    90M      \\ 
				& \checkmark         &  \checkmark   & \checkmark   & \textbf{57.1} & 28.3 & 2.02   &    90M      \\ \hline
				\multicolumn{8}{|c|}{Hyneter Plus}                                                                                                         \\ \hline
				\multirow{3}{*}{baseline} &  \checkmark         &     &    & 54.8 & 23.0 & 2.38   &      125M    \\ 
				& \checkmark          &  \checkmark   &    & 56.4 & 26.7 & 2.11   &     134M     \\ 
				& \checkmark         &   \checkmark  & \checkmark   & \textbf{58.0} & 27.9 & 2.08   &     134M     \\ \hline
				\multicolumn{8}{|c|}{Hyneter Max}                                                                                                         \\ \hline
				\multirow{3}{*}{baseline}  &  \checkmark         &     &    & 55.7 & 25.7 & 2.17   &   227M       \\ 
				&  \checkmark         & \checkmark    &    & 58.3 & 27.4 & 2.10   &    247M      \\  
				&  \checkmark         & \checkmark    & \checkmark   & \textbf{60.1} & 29.8 & 2.07   &     247M     \\ \hline
		\end{tabular}}
		\caption{Object detection performance (\%) on Hyneter vatiants with Mask R-CNN frameworks on MS COCO \texttt{test-dev} set. \textit{Originals} means pure Transformer baselines without HNB or DS, which is similar to Swin-T structurally.}
		\label{tab:abla}
	\end{table}
	
	HNB extends the range of local information, retaining and transforing local information and global dependencies to \textit{Neck} simultaneously, which greatly increases the proportion of small object performance ($AP/AP_{s}:$ 2.43 $\rightarrow$ 2.17 ;  2.38 $\rightarrow$ 2.11 ; 2.17 $\rightarrow$ 2.10), thereby improve general performance ($AP:$ 52.3 $\rightarrow$ 55.0 ;  54.8 $\rightarrow$ 56.4 ; 55.7 $\rightarrow$ 58.3). With the deepening of the stages, self-attention will constantly weaken local information and increase the role of global dependencies. Meanwhile, DS will retain local information in the patch, and restrain the excessive strengthening of existing global dependencies, which improve $AP$ and $AP_{s}$ concurrently (see Table \ref{tab:abla}).
	
	\subsection{Object Detection on MS COCO}
	
	\textbf{Setting.} For the ablation study, we consider 4 typical object detection frameworks: Mask R-CNN, ATSS, DETR, and Swin Transformer with the same setting (multi-scale training, ADamW optimizer with initial learning rate of 0.00001 and weight decay of 0.05 ) in mmdetection \cite{chen2019mmdetection}. We adopt ImageNet-22K pre-trained model as initialization for system-level comparison. 
	
	\textbf{Dataset} is mentioned in \textbf{Ablation studies}.
	
	\begin{table}[h]\small

		\setlength{\tabcolsep}{1.3mm}{
			\begin{tabular}{|l|l|c|c|c|c|}
				\hline
				Method                      & \multicolumn{1}{c|}{Backbone} & $AP$   & $AP_{s}$  & $AP/AP_{s}$ & \#param. \\ \hline
				\multirow{2}{*}{Mask R-CNN} & R-50                          & 42.3 & 24.7 & 1.71   & 82M      \\  
				& Hyneter-plus                  & 58.0 & 27.9 & 2.07   & 134M     \\ \hline
				\multirow{2}{*}{ATSS}       & R-50                          & 43.5 & 25.7 & 1.69   & 32M      \\ 
				& Hyneter-plus                  & 56.0 & 27.4 & 2.04   & 53M      \\ \hline
				\multirow{2}{*}{DETR}       & R-50 + trans                  & 42.0 & 20.5 & 2.05   & 41M      \\ 
				& Hyneter-plus                  & 47.0 & 24.7 & 1.90   & 93M      \\ \hline
		\end{tabular}}
		\caption{Object detection performance (\%) with various frameworks on MS COCO val set.  \textit{R50 + trans} means that R50 and Transformer Blocks as DETR Backbone.}
		\label{tab:obco1}
	\end{table}
	
	\begin{table}[h]\small

		\setlength{\tabcolsep}{3.1mm}{
			\begin{tabular}{|l|c|c|c|c|}
				\hline
				Backbone      & $AP$   & $AP_{s}$  & $AP/AP_{s}$  & \#params. \\ \hline
				R-50         & 42.3 & 24.7 & 1.71   & 82M       \\ 
				R-101        & 44.5 & 25.5 & 1.74   & 101M      \\ \hline
				Swin-T       & 49.8 & 21.4 & 2.33   & 86M       \\ 
				Swin-S       & 51.4 & 25.1 & 2.05   & 107M      \\ 
				Swin-B       & 51.5 & 25.0 & 2.06   & 145M      \\ 
				Swin-L       & 57.8 & 26.7 & 2.16   & 284M      \\ \hline
				Hyneter-1.0  & 57.1 & 28.3 & 2.02   & 90M       \\ 
				Hyneter-plus & 58.0 & 27.4 & 2.08   & 134M      \\ 
				Hyneter-Max  & \textbf{60.1} & \textbf{29.8} & 2.07   & 247M      \\ \hline
		\end{tabular}}
		\caption{Object detection (with Mask R-CNN) performance (\%) with various backbones on COCO val set.}
		\label{tab:obco2}
	\end{table}
	
	\begin{table}[h]\small

		\setlength{\tabcolsep}{1.7mm}{
			\begin{tabular}{|lllll|}
				\hline
				\multicolumn{1}{|l|}{Method}                 & \multicolumn{1}{c|}{AP}   & \multicolumn{1}{c|}{APs}  & \multicolumn{1}{c|}{AP/APs} & \multicolumn{1}{c|}{\#param.} \\ \hline
				\multicolumn{1}{|l|}{ATSS(ResNeXt-101-DCN)}  & \multicolumn{1}{l|}{50.7} & \multicolumn{1}{l|}{33.2} & \multicolumn{1}{l|}{1.53}   & --                            \\ 
				\multicolumn{1}{|l|}{EfficientDet-D7x(1537)} & \multicolumn{1}{l|}{55.1} & \multicolumn{1}{l|}{--}   & \multicolumn{1}{l|}{--}     & 77M                           \\ \hline
				\multicolumn{5}{|c|}{DETR series Backbone: DC5-R50 or R50}                                                                                                         \\ \hline
				\multicolumn{1}{|l|}{DETR}                   & \multicolumn{1}{l|}{43.3} & \multicolumn{1}{l|}{22.5} & \multicolumn{1}{l|}{1.92}   & 41M                           \\ 
				\multicolumn{1}{|l|}{UP-DETR }                & \multicolumn{1}{l|}{42.8} & \multicolumn{1}{l|}{20.8} & \multicolumn{1}{l|}{2.06}   & --                            \\ 
				\multicolumn{1}{|l|}{Deformable DETR}        & \multicolumn{1}{l|}{46.9} & \multicolumn{1}{l|}{27.7} & \multicolumn{1}{l|}{1.69}   & --                            \\ 
				\multicolumn{1}{|l|}{Conditional DETR}       & \multicolumn{1}{l|}{45.1} & \multicolumn{1}{l|}{25.3} & \multicolumn{1}{l|}{1.78}   & 44M                           \\ \hline
				\multicolumn{5}{|c|}{Swin Transformer with Cascade Mask R-CNN}                                                                                                     \\ \hline
				\multicolumn{1}{|l|}{Swin-B (HTC++)}        & \multicolumn{1}{l|}{56.4} & \multicolumn{1}{l|}{25.1}   & \multicolumn{1}{l|}{2.25}     & 160M                          \\
				\multicolumn{1}{|l|}{Swin-L (HTC++)}        & \multicolumn{1}{l|}{57.1} & \multicolumn{1}{l|}{25.6}   & \multicolumn{1}{l|}{2.23}     & 284M                          \\
				\multicolumn{1}{|l|}{Swin-L (HTC++)*}        & \multicolumn{1}{l|}{58.0} & \multicolumn{1}{l|}{26.0}   & \multicolumn{1}{l|}{2.23}     & 284M                          \\ \hline
				\multicolumn{5}{|c|}{Ours with Mask R-CNN}                                                                                                                         \\ \hline
				\multicolumn{1}{|l|}{Hyneter-1.0}            & \multicolumn{1}{l|}{57.1} & \multicolumn{1}{l|}{28.3} & \multicolumn{1}{l|}{2.02}   & 90M                           \\ 
				\multicolumn{1}{|l|}{Hyneter-plus}           & \multicolumn{1}{l|}{58.0} & \multicolumn{1}{l|}{27.9} & \multicolumn{1}{l|}{2.08}   & 134M                          \\ 
				\multicolumn{1}{|l|}{Hyneter-Max}            & \multicolumn{1}{l|}{\textbf{60.1}} & \multicolumn{1}{l|}{\textbf{29.8}} & \multicolumn{1}{l|}{2.07}   & 247M                          \\ \hline
		\end{tabular}}
		\caption{System-level comparison (\%) on MS COCO \texttt{test-dev} set.  * indicates multi-scale testing. The frameworks in Swin Trans (Swin-Transformer \cite{liu2021swin}) is Cascade Mask R-CNN. EfficientDet-D7x(1537)\cite{2020EfficientDet}}
		\label{tab:obco3}
	\end{table}
	
	\vspace{-0.2cm} 
	
	\textbf{Comparison to ResNet.} The results of Hyneter-plus and R-50 on 4 object detection frameworks are listed in Table \ref{tab:obco1}. Our Hyneter-plus architecture brings consistent $+ 5.0 \sim 15.7$ $AP$ and $ + 1.7 \sim 4.2$ $AP_{S}$ gains over ResNet-50, with acceptable larger model size. All Hyneters achieve significant gains of $+ 14.8 \sim 15.6 AP$ and $+ 3.6 \sim 4.3 AP_{S}$ over ResNet-50 or ResNet-101, which have similar or lighter model size (see Table \ref{tab:obco2}). 
	
	\textbf{Comparison to Swin Transformer.} The comparison of Hyneter and Swin Transformer under different backbones with Mask R-CNN is showed in Table \ref{tab:obco2}. Hyneters achieve a high detection accuracy of $60.1 AP$ and $29.8 AP_{S}$, which are significant improvement of $+ 2.3 \sim 7.3$ $AP$ and $ + 3.1 \sim 6.9 $ $AP_{S}$ over Swin series methods with lighter model size.
	
	\textbf{Comparison to previous state-of-the-art.} Table \ref{tab:obco3} lists the comparison of our best results with precious state-of-the-art methods. Hyneter method achieves $ + 60.1 AP$ and $ 29.8 AP_{S}$on COCO \textit{test-dev} set, surpassing the previous best performances by $ + 9.4  AP$ (ATSS \cite{zhang2020bridging}), $ + 5.0 AP$ (EfficientDet-D7x \cite{2020EfficientDet}), $ + 13.2 AP$ (Deformable DETR \cite{zhu2020deformable}), and $ + 2.1 AP$ (Swin-L \cite{liu2021swin} with HTC++ and multi-scale testing). Furthermore, Hyneters greatly improve $AP_{S}$, comparing with Swin Transformer series.
	
	\subsection{Object Detection on VisDrone}
	
	\textbf{Setting.} For comparison, we consider methods with the same setting (multi-scale training, AdamW optimizer with initial learning rate of 0.00001 and weight decay of 0.05 ) in mmdetection. We adopt ImageNet-22K pre-trained model as initialization for system-level comparison.
	
	\textbf{Dataset.} The Visdrone dataset consists of 400 video clips formed by 265,228 frames and 10,209 static images, captured by various drone-mounted cameras, covering a wide range of aspects including location, environment, objects (10 classes). These frames are manually annotated with more than 2.6 million bounding boxes or points of targets of frequent interests, such as pedestrians, cars, bicycles, and tricycles.
	
	\begin{table}[h]\small
		
		\setlength{\tabcolsep}{5.2mm}{
			\begin{tabular}{|l|c|c|c|}
				\hline
				Method                       & $AP$ & $AP_{50}$ & $AP_{75}$ \\ \hline
				DBNet &39.4 &65.4 &41.0     \\ 
				SOLOer  &39.4 &63.9& 40.8   \\ 
				Swin-T& 39.4 &63.9 &40.8     \\ 
				TPH-YOLOv5 &39.1 &62.8 &41.3     \\ 
				VistrongerDet &38.7 &64.2 &40.2      \\ 
				EfficientDet &38.5 &63.2 &39.5     \\ 
				DroneEye2020 &34.5 &58.2 &35.7     \\ 
				Cascade R-CNN &16.0 &31.9 &15.0      \\ 
				DPNet-ensemble &37.3 &62.0 &39.1     \\ \hline
				Hyneter 1.0& 41.9   &   65.8   &  43.7    \\ 
				Hyneter plus& 43.7   &   70.1   &  45.8    \\ 
				Hyneter Max & \textbf{46.1}   &   \textbf{73.9}   &  \textbf{47.0}   \\ \hline
		\end{tabular}}
		\caption{System-level comparison of Hyneters with Mask R-CNN performance (\%) on VisDrone-DET2021 \cite{cao2021visdrone}. }
		\label{tab:obvis}
	\end{table}
	
	Table \ref{tab:obvis} compares our best results with those of previous state-of-the-art models on the VisDrone-DET2021 Challenge \cite{cao2021visdrone} . Our best model (Hyneter Max) achieves $ 46.1  AP,  73.9  AP_{50} $, and $ 47.0 AP_{75} $ on the VisDrone, surpassing all previous best results in Table \ref{tab:obvis}.
	
	\subsection{Instance Segmentation on MS COCO}
	
	\textbf{Setting} and \textbf{Dataset} are mentioned in \textbf{Ablation studies}.
	
	\begin{table}[h]\small
		
		\setlength{\tabcolsep}{2.1mm}{
			\begin{tabular}{|l|c|c|c|c|}
				\hline
				Method          & $AP^{mask}$   & $AP^{mask}_{50}$  & $AP^{mask}_{75}$  & \#param. \\ \hline
				R-50              & 32.5 & 55.4 & 31.7   & 82M       \\ 
				R-101             & 35.9 & 60.7 & 36.8   & 101M      \\ \hline
				Swin-T           & 40.0 & 68.7 & 42.3   & 86M       \\ 
				Swin-S           & 41.5 & 70.1 & 42.0   & 107M      \\ 
				Swin-B           & 42.0 & 74.0 & 42.6   & 145M      \\ \hline
				Hyneter-1.0    & 45.1 & 78.3 & 42.2   & 90M       \\ 
				Hyneter-plus  & 46.9 & 79.9 & 45.0   & 134M      \\ 
				Hyneter-Max  & \textbf{48.5} & \textbf{82.1} & \textbf{46.7}   & 247M      \\ \hline
		\end{tabular}}
		\caption{Instance segmentation (with Mask R-CNN) performance (\%) with various backbones  on MS COCO \texttt{test-dev} set. }
		\label{tab:isco}
	\end{table}
	
	Table \ref{tab:isco} compares our best instance segmentation results with those of previous state-of- the-art models on COCO. Our best model (Hyneter Max) achieves $ 48.5 AP^{mask}$,  $82.1  AP^{mask}_{50}$ , and  $46.7 AP^{mask}_{75}$ with competitive model size, surpassing all previous best results (see Table \ref{tab:isco}).
	
	\subsection{Semantic Segmentation on ADE20K}
	
	\textbf{Setting.} In training, we employ the AdamW optimizer with an initial learning rate of $1.0 \times 10^{-5} $, a weight decay of 0.01, a scheduler that uses linear learning rate decay, and a linear warmup of 1,500 iterations. Models are trained on 2 GPUs with 4 images per GPU for 140K iterations
	
	\textbf{Dataset.} ADE20K has more than 25K images of complex daily scenes, including various objects in natural space environment (20.2k for training, 2K for validation, 3K for test). ADE20K covers various annotations of scenes, objects and object parts, and each image has an average of 19.5 instances and 10.5 object classes. 
	
	\begin{table}[h]\small
		
		\setlength{\tabcolsep}{2.1mm}{
			\begin{tabular}{|l|c|c|c|c|}
				\hline
				Method   & Backbone     & val mIoU & test score & \#param. \\ \hline
				DANet    & ResNet-101   & 45.2     & --         & 69M      \\ 
				Dlab.v3+ & ResNet-101   & 44.1     & --         & 63M      \\ 
				OCRNet   & ResNet-101   & 45.3     & 56.0       & 56M      \\ 
				UperNet  & ResNet-101   & 44.9     & --         & 86M      \\ \hline
				OCRNet   & HRNet-w48    & 45.7     & --         & 71M      \\ 
				Dlab.v3+ & ResNeSt-101  & 46.9     & 55.1       & 66M      \\ 
				Dlab.v3+ & ResNeSt-200  & 48.4     & --         & 88M      \\ 
				SETR     & T-Large      & 50.3     & 61.7       & 308M     \\ \hline
				UperNet  & Swin-S       & 49.3     & --         & 81M      \\ 
				UperNet  & Swin-B       & 51.6     & --         & 121M     \\ 
				UperNet  & Swin-L       & 53.5     & 62.8       & 234M     \\ \hline
				UperNet  & Hyneter 1.0  & 50.6     & 62.0       & 82M      \\ 
				UperNet  & Hyneter Plus & 53.0     & 63.4       & 125M     \\ 
				UperNet  & Hyneter Max  & \textbf{54.3}     & \textbf{65.9}       & 231M     \\ \hline
		\end{tabular}}
		\caption{Results of semantic segmentation on the ADE20K val and test set. The comparison data is from Appendix A2.3 in \cite{liu2021swin}.}
		\label{tab:ssADK}
	\end{table}
	
	Table \ref{tab:ssADK} lists the mIoU, and model size (\#param) for different method/backbone pairs. From these results, it can be seen that Hyneter Max is $+4.3 mIoU$ higher  than SETR  with much lighter model size. It is also $+6.0 mIoU$ higher than ResNeS200, and $+9.4 mIoU$ higher than ResNeSt-101. Our Hyneter series with UperNet achieve $50.6 mIoU, 53.0 mIoU$, and $54.3 mIoU $ on the val set, surpassing the previous Swin Transformer series by $ +0.8 \sim 1.4 mIoU$.
	
	\section{Conclusion}
	
	In this work, we point out that the essential differences between CNN-based and Transformer-based detectors are the gap between local information and global dependencies in feature extraction and propagation.  To address these differences, we propose a new vision Transformer, called \textbf{H}ybrid \textbf{N}etwork \textbf{T}ransformer (Hyneter), which consists of Hybrid Network Backbone (HNB) and Dual Switching module (DS). Based on the balance strategy, Hyneters integrate and transfer local information and global dependencies simultaneously, so they are able to significantly improve performance. Ablation studies illustrate that Hyneters with HNB and DS achieve the state-of-the-art performance on multiple datasets for object detection. Furthermore,  Hyneters achieve the state-of-the-art performance on multiple tasks (object detection,  semantic segmentation, and instance segmentation) significantly, and surpass previous best methods. We do hope that Hynerters will play a role of cornerstone to encourage balancing methods between local information and global dependencies in computer vision.
	
	
	

\vfill\pagebreak

\label{sec:refs}

\bibliographystyle{IEEEbib}
\bibliography{aaai23}

\begin{thebibliography}{10}

\bibitem{dong2021attention}
Yihe Dong, Jean-Baptiste Cordonnier, and Andreas Loukas,
\newblock ``Attention is not all you need: Pure attention loses rank doubly
  exponentially with depth,''
\newblock in {\em International Conference on Machine Learning}. PMLR, 2021,
  pp. 2793--2803.

\bibitem{chen2021you}
Qiang Chen, Yingming Wang, Tong Yang, Xiangyu Zhang, Jian Cheng, and Jian Sun,
\newblock ``You only look one-level feature,''
\newblock in {\em Proceedings of the IEEE/CVF conference on computer vision and
  pattern recognition}, 2021, pp. 13039--13048.

\bibitem{raghu2021vision}
Maithra Raghu, Thomas Unterthiner, Simon Kornblith, Chiyuan Zhang, and Alexey
  Dosovitskiy,
\newblock ``Do vision transformers see like convolutional neural networks?,''
\newblock {\em Advances in Neural Information Processing Systems}, vol. 34, pp.
  12116--12128, 2021.

\bibitem{li2022exploring}
Yanghao Li, Hanzi Mao, Ross Girshick, and Kaiming He,
\newblock ``Exploring plain vision transformer backbones for object
  detection,''
\newblock {\em arXiv preprint arXiv:2203.16527}, 2022.

\bibitem{dai2021dynamic}
Xiyang Dai, Yinpeng Chen, Bin Xiao, Dongdong Chen, Mengchen Liu, Lu~Yuan, and
  Lei Zhang,
\newblock ``Dynamic head: Unifying object detection heads with attentions,''
\newblock in {\em Proceedings of the IEEE/CVF conference on computer vision and
  pattern recognition}, 2021, pp. 7373--7382.

\bibitem{1998Gradient}
Y~Lecun and L~Bottou,
\newblock ``Gradient-based learning applied to document recognition,''
\newblock {\em Proceedings of the IEEE}, vol. 86, no. 11, pp. 2278--2324, 1998.

\bibitem{krizhevsky2017imagenet}
Alex Krizhevsky, Ilya Sutskever, and Geoffrey~E Hinton,
\newblock ``Imagenet classification with deep convolutional neural networks,''
\newblock {\em Communications of the ACM}, vol. 60, no. 6, pp. 84--90, 2017.

\bibitem{simonyan2014very}
Karen Simonyan and Andrew Zisserman,
\newblock ``Very deep convolutional networks for large-scale image
  recognition,''
\newblock {\em arXiv preprint arXiv:1409.1556}, 2014.

\bibitem{2014Going}
Christian Szegedy, Wei Liu, Yangqing Jia, Pierre Sermanet, and Andrew
  Rabinovich,
\newblock ``Going deeper with convolutions,''
\newblock 2014.

\bibitem{2016Deep}
Kaiming He, Xiangyu Zhang, Shaoqing Ren, and Jian Sun,
\newblock ``Deep residual learning for image recognition,''
\newblock in {\em IEEE Conference on Computer Vision \& Pattern Recognition},
  2016.

\bibitem{howard2017mobilenets}
Andrew~G Howard, Menglong Zhu, Bo~Chen, Dmitry Kalenichenko, Weijun Wang,
  Tobias Weyand, Marco Andreetto, and Hartwig Adam,
\newblock ``Mobilenets: Efficient convolutional neural networks for mobile
  vision applications,''
\newblock {\em arXiv preprint arXiv:1704.04861}, 2017.

\bibitem{tan2019efficientnet}
Mingxing Tan and Quoc~V Le,
\newblock ``Efficientnet: Rethinking model scaling for convolutional neural
  networks,''
\newblock {\em arXiv preprint arXiv:1905.11946}, 2019.

\bibitem{fu2017dssd}
Cheng-Yang Fu, Wei Liu, Ananth Ranga, Ambrish Tyagi, and Alexander~C Berg,
\newblock ``Dssd: Deconvolutional single shot detector,''
\newblock {\em arXiv preprint arXiv:1701.06659}, 2017.

\bibitem{noh2015learning}
Hyeonwoo Noh, Seunghoon Hong, and Bohyung Han,
\newblock ``Learning deconvolution network for semantic segmentation,''
\newblock in {\em Proceedings of the IEEE international conference on computer
  vision}, 2015, pp. 1520--1528.

\bibitem{radford2015unsupervised}
Alec Radford, Luke Metz, and Soumith Chintala,
\newblock ``Unsupervised representation learning with deep convolutional
  generative adversarial networks,''
\newblock {\em arXiv preprint arXiv:1511.06434}, 2015.

\bibitem{dai2017deformable}
Jifeng Dai, Haozhi Qi, Yuwen Xiong, Yi~Li, Guodong Zhang, Han Hu, and Yichen
  Wei,
\newblock ``Deformable convolutional networks,''
\newblock in {\em Proceedings of the IEEE international conference on computer
  vision(ICCV)}, 2017, pp. 764--773.

\bibitem{2019Deformable}
Xizhou Zhu, Han Hu, Stephen Lin, and Jifeng Dai,
\newblock ``Deformable convnets v2: More deformable, better results,''
\newblock in {\em 2019 IEEE/CVF Conference on Computer Vision and Pattern
  Recognition (CVPR)}, 2019.

\bibitem{chen2017rethinking}
Liang-Chieh Chen, George Papandreou, Florian Schroff, and Hartwig Adam,
\newblock ``Rethinking atrous convolution for semantic image segmentation,''
\newblock {\em arXiv preprint arXiv:1706.05587}, 2017.

\bibitem{2016Multi}
Fisher Yu and Vladlen Koltun,
\newblock ``Multi-scale context aggregation by dilated convolutions,''
\newblock 2016.

\bibitem{dosovitskiy2020image}
Alexey Dosovitskiy, Lucas Beyer, Alexander Kolesnikov, Dirk Weissenborn,
  Xiaohua Zhai, Thomas Unterthiner, Mostafa Dehghani, Matthias Minderer, Georg
  Heigold, Sylvain Gelly, et~al.,
\newblock ``An image is worth 16x16 words: Transformers for image recognition
  at scale,''
\newblock {\em arXiv preprint arXiv:2010.11929}, 2020.

\bibitem{touvron2021training}
Hugo Touvron, Matthieu Cord, Matthijs Douze, Francisco Massa, Alexandre
  Sablayrolles, and Herv{\'e} J{\'e}gou,
\newblock ``Training data-efficient image transformers \& distillation through
  attention,''
\newblock in {\em International Conference on Machine Learning}. PMLR, 2021,
  pp. 10347--10357.

\bibitem{yuan2021tokens}
Li~Yuan, Yunpeng Chen, Tao Wang, Weihao Yu, Yujun Shi, Zi-Hang Jiang,
  Francis~EH Tay, Jiashi Feng, and Shuicheng Yan,
\newblock ``Tokens-to-token vit: Training vision transformers from scratch on
  imagenet,''
\newblock in {\em Proceedings of the IEEE/CVF International Conference on
  Computer Vision}, 2021, pp. 558--567.

\bibitem{han2021Transformer}
Kai Han, An~Xiao, Enhua Wu, Jianyuan Guo, Chunjing Xu, and Yunhe Wang,
\newblock ``Transformer in transformer,''
\newblock {\em Advances in Neural Information Processing Systems}, vol. 34, pp.
  15908--15919, 2021.

\bibitem{wang2021pyramid}
Wenhai Wang, Enze Xie, Xiang Li, Deng-Ping Fan, Kaitao Song, Ding Liang, Tong
  Lu, Ping Luo, and Ling Shao,
\newblock ``Pyramid vision transformer: A versatile backbone for dense
  prediction without convolutions,''
\newblock in {\em Proceedings of the IEEE/CVF International Conference on
  Computer Vision}, 2021, pp. 568--578.

\bibitem{carion2020end}
Nicolas Carion, Francisco Massa, Gabriel Synnaeve, Nicolas Usunier, Alexander
  Kirillov, and Sergey Zagoruyko,
\newblock ``End-to-end object detection with transformers,''
\newblock in {\em European conference on computer vision}. Springer, 2020, pp.
  213--229.

\bibitem{liu2021swin}
Ze~Liu, Yutong Lin, Yue Cao, Han Hu, Yixuan Wei, Zheng Zhang, Stephen Lin, and
  Baining Guo,
\newblock ``Swin transformer: Hierarchical vision transformer using shifted
  windows,''
\newblock in {\em Proceedings of the IEEE/CVF International Conference on
  Computer Vision}, 2021, pp. 10012--10022.

\bibitem{dai2021up}
Zhigang Dai, Bolun Cai, Yugeng Lin, and Junying Chen,
\newblock ``Up-detr: Unsupervised pre-training for object detection with
  transformers,''
\newblock in {\em Proceedings of the IEEE/CVF conference on computer vision and
  pattern recognition}, 2021, pp. 1601--1610.

\bibitem{meng2021conditional}
Depu Meng, Xiaokang Chen, Zejia Fan, Gang Zeng, Houqiang Li, Yuhui Yuan, Lei
  Sun, and Jingdong Wang,
\newblock ``Conditional detr for fast training convergence,''
\newblock in {\em Proceedings of the IEEE/CVF International Conference on
  Computer Vision}, 2021, pp. 3651--3660.

\bibitem{gupta2022ow}
Akshita Gupta, Sanath Narayan, KJ~Joseph, Salman Khan, Fahad~Shahbaz Khan, and
  Mubarak Shah,
\newblock ``Ow-detr: Open-world detection transformer,''
\newblock in {\em Proceedings of the IEEE/CVF Conference on Computer Vision and
  Pattern Recognition}, 2022, pp. 9235--9244.

\bibitem{zhu2020deformable}
Xizhou Zhu, Weijie Su, Lewei Lu, Bin Li, Xiaogang Wang, and Jifeng Dai,
\newblock ``Deformable detr: Deformable transformers for end-to-end object
  detection,''
\newblock {\em arXiv preprint arXiv:2010.04159}, 2020.

\bibitem{srinivas2021bottleneck}
Aravind Srinivas, Tsung-Yi Lin, Niki Parmar, Jonathon Shlens, Pieter Abbeel,
  and Ashish Vaswani,
\newblock ``Bottleneck transformers for visual recognition,''
\newblock in {\em Proceedings of the IEEE/CVF conference on computer vision and
  pattern recognition}, 2021, pp. 16519--16529.

\bibitem{ramachandran2019stand}
Prajit Ramachandran, Niki Parmar, Ashish Vaswani, Irwan Bello, Anselm Levskaya,
  and Jon Shlens,
\newblock ``Stand-alone self-attention in vision models,''
\newblock {\em Advances in Neural Information Processing Systems}, vol. 32,
  2019.

\bibitem{hu2019local}
Han Hu, Zheng Zhang, Zhenda Xie, and Stephen Lin,
\newblock ``Local relation networks for image recognition,''
\newblock in {\em Proceedings of the IEEE/CVF International Conference on
  Computer Vision}, 2019, pp. 3464--3473.

\bibitem{zhao2020exploring}
Hengshuang Zhao, Jiaya Jia, and Vladlen Koltun,
\newblock ``Exploring self-attention for image recognition,''
\newblock in {\em Proceedings of the IEEE/CVF Conference on Computer Vision and
  Pattern Recognition}, 2020, pp. 10076--10085.

\bibitem{wang2020axial}
Huiyu Wang, Yukun Zhu, Bradley Green, Hartwig Adam, Alan Yuille, and
  Liang-Chieh Chen,
\newblock ``Axial-deeplab: Stand-alone axial-attention for panoptic
  segmentation,''
\newblock in {\em European Conference on Computer Vision}. Springer, 2020, pp.
  108--126.

\bibitem{ho2019axial}
Jonathan Ho, Nal Kalchbrenner, Dirk Weissenborn, and Tim Salimans,
\newblock ``Axial attention in multidimensional transformers,''
\newblock {\em arXiv preprint arXiv:1912.12180}, 2019.

\bibitem{sun2019videobert}
Chen Sun, Austin Myers, Carl Vondrick, Kevin Murphy, and Cordelia Schmid,
\newblock ``Videobert: A joint model for video and language representation
  learning,''
\newblock in {\em Proceedings of the IEEE/CVF International Conference on
  Computer Vision}, 2019, pp. 7464--7473.

\bibitem{lu2019vilbert}
Jiasen Lu, Dhruv Batra, Devi Parikh, and Stefan Lee,
\newblock ``Vilbert: Pretraining task-agnostic visiolinguistic representations
  for vision-and-language tasks,''
\newblock {\em Advances in neural information processing systems}, vol. 32,
  2019.

\bibitem{huang2019ccnet}
Zilong Huang, Xinggang Wang, Lichao Huang, Chang Huang, Yunchao Wei, and Wenyu
  Liu,
\newblock ``Ccnet: Criss-cross attention for semantic segmentation,''
\newblock in {\em Proceedings of the IEEE/CVF international conference on
  computer vision}, 2019, pp. 603--612.

\bibitem{bello2019attention}
Irwan Bello, Barret Zoph, Ashish Vaswani, Jonathon Shlens, and Quoc~V Le,
\newblock ``Attention augmented convolutional networks,''
\newblock in {\em Proceedings of the IEEE/CVF international conference on
  computer vision}, 2019, pp. 3286--3295.

\bibitem{vaswani2017attention}
Ashish Vaswani, Noam Shazeer, Niki Parmar, Jakob Uszkoreit, Llion Jones,
  Aidan~N Gomez, {\L}ukasz Kaiser, and Illia Polosukhin,
\newblock ``Attention is all you need,''
\newblock {\em Advances in neural information processing systems}, vol. 30,
  2017.

\bibitem{liu2022swin}
Ze~Liu, Han Hu, Yutong Lin, Zhuliang Yao, Zhenda Xie, Yixuan Wei, Jia Ning, Yue
  Cao, Zheng Zhang, Li~Dong, et~al.,
\newblock ``Swin transformer v2: Scaling up capacity and resolution,''
\newblock in {\em Proceedings of the IEEE/CVF Conference on Computer Vision and
  Pattern Recognition}, 2022, pp. 12009--12019.

\bibitem{loshchilov2017decoupled}
Ilya Loshchilov and Frank Hutter,
\newblock ``Decoupled weight decay regularization,''
\newblock {\em arXiv preprint arXiv:1711.05101}, 2019.

\bibitem{chen2019mmdetection}
Kai Chen, Jiaqi Wang, Jiangmiao Pang, Yuhang Cao, Yu~Xiong, Xiaoxiao Li,
  Shuyang Sun, Wansen Feng, Ziwei Liu, Jiarui Xu, et~al.,
\newblock ``Mmdetection: Open mmlab detection toolbox and benchmark,''
\newblock {\em arXiv preprint arXiv:1906.07155}, 2019.

\bibitem{2020EfficientDet}
Mingxing Tan, Ruoming Pang, and Quoc~V. Le,
\newblock ``Efficientdet: Scalable and efficient object detection,''
\newblock in {\em 2020 IEEE/CVF Conference on Computer Vision and Pattern
  Recognition (CVPR)}, 2020.

\bibitem{zhang2020bridging}
Shifeng Zhang, Cheng Chi, Yongqiang Yao, Zhen Lei, and Stan~Z. Li,
\newblock ``Bridging the gap between anchor-based and anchor-free detection via
  adaptive training sample selection,''
\newblock in {\em CVPR}, 2020.

\bibitem{cao2021visdrone}
Yaru Cao, Zhijian He, Lujia Wang, Wenguan Wang, Yixuan Yuan, Dingwen Zhang,
  Jinglin Zhang, Pengfei Zhu, Luc Van~Gool, Junwei Han, et~al.,
\newblock ``Visdrone-det2021: The vision meets drone object detection challenge
  results,''
\newblock in {\em Proceedings of the IEEE/CVF International Conference on
  Computer Vision}, 2021, pp. 2847--2854.

\end{thebibliography}

\end{document}